%% file: acl_latex.tex
\pdfoutput=1

\documentclass[11pt]{article}

\usepackage[final]{acl}

\usepackage{times}
\usepackage{latexsym}

\usepackage{booktabs}
\usepackage{multirow}
\usepackage{graphicx} 
\usepackage{amsmath}  
\usepackage{hyperref} 
\usepackage{adjustbox}
\usepackage{enumitem}
\usepackage{algorithm}
\usepackage{algorithmic}
\usepackage{listings}

\usepackage{titlesec}
\titlespacing*{\subsection}{0pt}{1.0ex plus 1ex minus .2ex}{1.0ex plus .2ex}

\lstdefinestyle{json}{
    basicstyle=\small\ttfamily,
    showstringspaces=false,
    breaklines=true,
    frame=none, 
    aboveskip=0pt, 
    belowskip=0pt, 
    xleftmargin=0pt,
    xrightmargin=0pt,
    backgroundcolor=\color{gray!10}, 
    numbers=left,
    numberstyle=\tiny,
    nolol=true
}
\usepackage{caption}
\usepackage{subcaption}
\usepackage{tcolorbox}
\usepackage{color}
\usepackage{colortbl}
\newcommand{\ours}{\textsc{IEPile}}

\definecolor{aliceblue}{RGB}{178, 217, 245}
\newcommand{\CC}{\cellcolor{aliceblue}}
\definecolor{babyblue}{RGB}{217, 239, 251}
\newcommand{\CB}{\cellcolor{babyblue}}
\usepackage{pifont}

\definecolor{my_green}{RGB}{51,102,0}
\definecolor{my_red}{RGB}{204, 0, 0}

\definecolor{c3}{cmyk}{0.9081,0,0.7209,0.5255}

\newtcbox{\hlprimarytab}{on line, rounded corners, box align=base, colback=c3!10,colframe=white,size=fbox,arc=3pt, before upper=\strut, top=-2pt, bottom=-4pt, left=-2pt, right=-2pt, boxrule=0pt}
\newtcbox{\hlsecondarytab}{on line, box align=base, colback=red!10,colframe=white,size=fbox,arc=3pt, before upper=\strut, top=-2pt, bottom=-4pt, left=-2pt, right=-2pt, boxrule=0pt}

\newlength\myheight
\newlength\mydepth
\usepackage[T1]{fontenc}

\usepackage[utf8]{inputenc}

\usepackage{microtype}

\usepackage{inconsolata}

\title{\textsc{IEPile}: Unearthing Large-Scale Schema-Based\\Information Extraction Corpus}

\author{
    Honghao Gui$^\spadesuit$$^\diamondsuit$\thanks{$\quad$ Equal Contribution.},
    Lin Yuan$^\clubsuit$$^\diamondsuit$\footnotemark[1], 
    Hongbin Ye$^\spadesuit$,
    \textbf{Ningyu Zhang}$^\spadesuit$$^\diamondsuit$\thanks{$\quad$ Corresponding Author.}\\
    \textbf{Mengshu Sun}$^\clubsuit$$^\diamondsuit$,
    \textbf{Lei Liang}$^\clubsuit$$^\diamondsuit$,
    \textbf{Huajun Chen}$^\spadesuit$$^\diamondsuit$\footnotemark[2]\\
    $^\spadesuit$ Zhejiang University 
    $^\diamondsuit$ Zhejiang University - Ant Group Joint Laboratory of Knowledge Graph\\
    $^\clubsuit$ Ant Group
    \texttt{
    \{guihonghao,zhangningyu,huajunsir\}@zju.edu.cn 
    }\\
  \url{https://github.com/zjunlp/IEPile}
}

\begin{document}
\maketitle
\begin{abstract}
Large Language Models (LLMs) demonstrate remarkable potential across various domains; however, they exhibit a significant performance gap in Information Extraction (IE). Note that high-quality instruction data is the vital key for enhancing the specific capabilities of LLMs, while current IE datasets tend to be small in scale, fragmented, and lack standardized schema. To this end, we introduce {\ours}, a comprehensive bilingual (English and Chinese) IE instruction corpus, which contains approximately \textbf{0.32B} tokens. We construct {\ours} by collecting and cleaning 33 existing IE datasets, and introduce schema-based instruction generation to unearth a large-scale corpus. Experimentally, {\ours} enhance the performance of LLMs for IE, with notable improvements in zero-shot generalization. We open-source the resource and pre-trained models, hoping to provide valuable support to the NLP community.
\end{abstract}

\section{Introduction}
Large Language Models (LLMs) have achieved significant breakthroughs in multiple Natural Language Processing (NLP) tasks \cite{du2022glm,DBLP:journals/corr/abs-2307-09288,DBLP:journals/corr/abs-2310-06825,DBLP:journals/corr/abs-2303-18223,DBLP:journals/corr/abs-2309-09558,DBLP:journals/corr/abs-2401-00812,DBLP:conf/bigdataconf/WuGCWY23,DBLP:journals/corr/abs-2312-17120,DBLP:journals/corr/abs-2401-14624}.
However, recent studies ~\citep{DBLP:journals/corr/abs-2304-11633,DBLP:conf/emnlp/Ma0HS23,DBLP:journals/corr/abs-2312-17617,DBLP:conf/acl/WadhwaAW23,DBLP:conf/emnlp/WanCMLSLK23,DBLP:journals/corr/abs-2310-05092,DBLP:conf/acl/LiSTYWHQ23,DBLP:conf/emnlp/Jiao0LZOJ023,DBLP:journals/corr/abs-2311-09562,wang2024techgpt20} indicate a significant performance gap in the task of Information Extraction (IE) when utilizing LLMs. 
~\citep{DBLP:conf/acl/LeeINZECC22,DBLP:journals/corr/abs-2310-05092} further illustrate that the major reason may lie in limited high-quality, large-scale data corpus. 
Concretely, most IE datasets are often limited in size, scattered in distribution, and lack standardization in schema\footnote{We refer to the schema as pre-defined types of entities, relations, events (arguments and roles), etc.}.

Faced with these limitations, there is an urgent need to collect instruction data in a unified and automated manner to build a high-quality, large-scale IE corpus.
To this end, we collect and clean various existing IE datasets to obtain a comprehensive bilingual IE instruction dataset named {\ours}\footnote{{\ours} adhere to the CC BY-NC-SA 4.0 license except for ACE2005 which adheres to the LDC User Agreement.}. 
During the corpus construction, we find existing methods for constructing IE instruction data suffer from two issues for generalizable IE: 
1) \textbf{Schema Query Disparity}: There may be inconsistency in the number of schema queries within instruction between training and evaluation which can harm model generalization;
2) \textbf{Semantic Confusion}: The co-occurrence of semantically similar schemas within instructions may confuse the model.
Thus, we introduce a schema-based instruction generation strategy.
We first construct a hard negative schema dictionary to promote the more frequent occurrence of semantically similar schema in instructions. 
Then, we introduce batched instruction generation, dynamically limiting the number of schemas queried in each instruction to $split\_num$, which not only addresses the issue of performance degradation due to inconsistent numbers of schema queries during training and evaluation, but also enhances the robustness when dealing with semantically confusing schema.
Finally, we obtain {\ours} which contains approximately 0.32B tokens.

By fine-tuning a selection of the latest prominent models~\citep{DBLP:journals/corr/abs-2309-10305,DBLP:journals/corr/abs-2307-09288,qwen} on the {\ours} dataset, we show that LLMs with {\ours} can yield better zero-shot performance than baselines.
This achievement not only verifies the effectiveness of the {\ours} dataset but also provides a framework for creating IE datasets in other domains.

\input{iepile}

\input{experiments}

\section{Conclusion and Future Work}
In this paper, we introduce {\ours}, by collecting and cleaning existing IE datasets and utilizing a schema-based instruction generation strategy. 
In the future, we will continue to maintain the corpus and try to integrate new resources including open-domain IE, and document-level IE.

\section*{Limitations}
\label{adx:limitation}
From the data perspective, our study primarily focuses on schema-based IE, which limits our ability to generalize to human instructions that do not follow our specific format requirements. 
Additionally, we do not explore the field of Open Information Extraction (Open IE); however, if we remove schema constraints, our dataset would be suitable for Open IE scenarios.
Besides,  {\ours} is confined to data in English and Chinese, and in the future, we hope to include data in more languages.
From the model's perspective, our research evaluates limited models, along with a few baselines due to the computation resources. 
Theoretically, {\ours} can be applied to any other LLMs such as ChatGLM \cite{du2022glm} and Gemma \footnote{\url{https://huggingface.co/google/gemma-7b}.}.

\section*{Ethical Considerations}
In this paper, we strictly adhered to the standards and principles of ethics. 
All data collected are sourced from publicly available materials, ensuring the transparency and legality of the research. 
We conduct a thorough review of the data used, verifying the legitimacy of their sources and compliance with their usage, thus avoiding any infringement on personal privacy or involvement with unauthorized information.

\section*{Acknowledgements}
We would like to express our sincere gratitude to the anonymous reviewers for their thoughtful and constructive feedback.
This work was supported by the National Natural Science Foundation of China (No. 62206246), the Fundamental Research Funds for the Central Universities (226-2023-00138), Zhejiang Provincial Natural Science Foundation of China (No. LGG22F030011), Yongjiang Talent Introduction Programme (2021A-156-G), CCF-Baidu Open Fund, and Information Technology Center and State Key Lab of CAD\&CG, Zhejiang University.

\bibliography{custom}

\input{appendix}

\end{document}

%% file: iepile.tex
\begin{figure*}[!t]
\begin{center}
\resizebox{0.99\textwidth}{!}{
\includegraphics{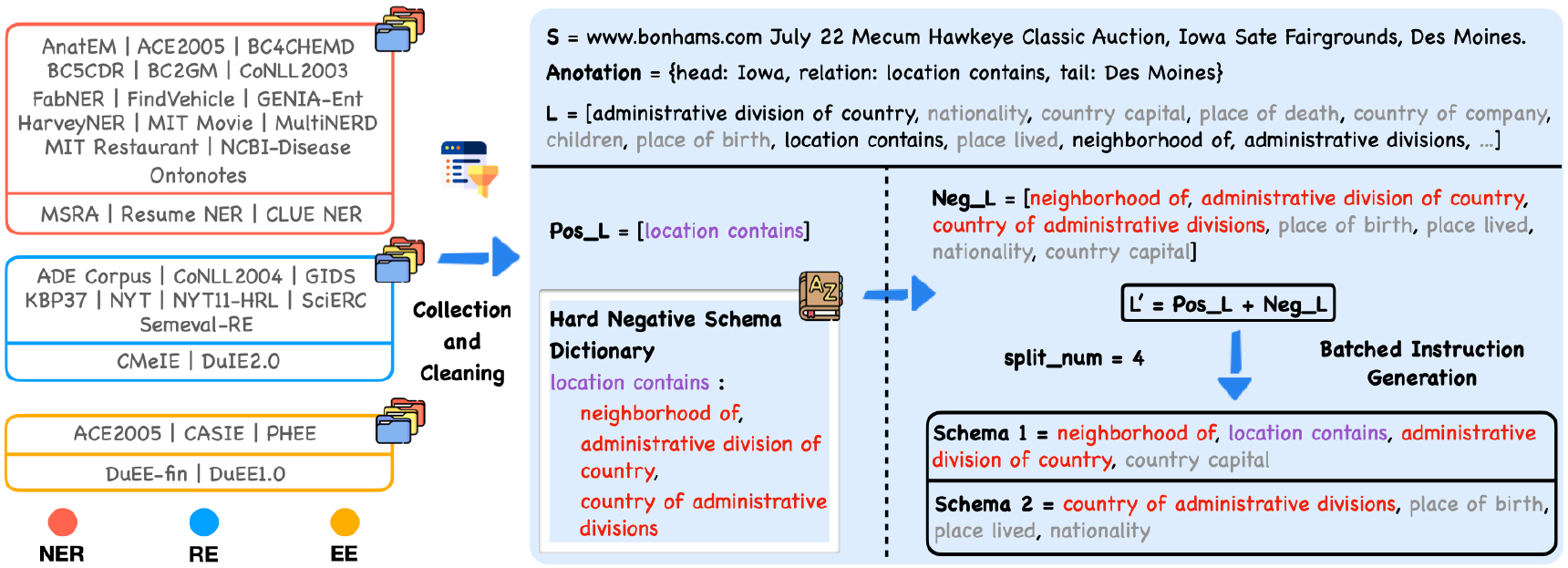}
}
\caption{An overview of the construction of {\ours}, including Data Collection and Cleaning, as well as Schema-Based Instruction Generation (Hard Negative Schema Construction and Batched Instruction Generation).}
\label{fig:dataconstruct}
\end{center}
\end{figure*}

\section{\textsc{IEPile}}
In this section, we introduce the construction of \textsc{IEPile} and provide details in Appendix ~\ref{apd:iepile}.

\subsection{Data Collection and Cleaning}
To broadly cover various domains and meet the practical demands, we collect datasets necessary for IE from multiple data sources. 
Our corpus mainly involves bilingual data (Chinese and English) and focuses on three principal categories of IE tasks: Named Entity Recognition (NER), Relation Extraction (RE), and Event Extraction (EE). 
In total, we gather 26 English datasets and 7 Chinese datasets. 
We also employ standardization procedures to maintain data quality and format uniformity, involving format unification, instance deduplication, and the exclusion of low-quality data.

\subsection{Schema-Based Instruction Generation}

We concentrate on instruction-based information extraction (IE), a methodology that incorporates three crucial elements to compose an instruction: 1) \textbf{Task Description}, a template utilized to distinguish between different IE tasks; 2) \textbf{Input Text}, the source text to be extracted; and 3) \textbf{Schema sequence}, which defines the information that the model is supposed to extract, including entity types, relations, events, etc. 
Among these, the schema sequence is critical as it reflects the specific extraction requirements and is dynamically variable. 
Therefore, the construction of the schema sequence within an instruction holds critical significance.

\paragraph{Positive and Negative Schema Mechanism in Instructions.} 
Firstly, we define schemas that actually exist within the input text as \textbf{positive schemas} and those that do not appear as \textbf{negative schemas}. 
As illustrated in Figure \ref{fig:dataconstruct}, the ``location contains'' present in the annotation is a positive schema, while all other schemas from the predefined label set $L$ are negative schemas.
Traditional IE frameworks, which are treated as sequence labeling tasks, take text as input and produce a label for each token as output, without involving the concept of positive or negative schemas within the model's input. 
However, in the era of generative IE, represented by models like UIE ~\citep{DBLP:conf/acl/0001LDXLHSW22}, introduce the concept of integrating a schema sequence (refers to as Structural Schema Instructor, or SSI) in the model's input to guide its output, restricting the range of output to the SSI.
The method necessitates including the entire predefined label set of a dataset as the SSI to guide the model's output during inference.
As a result, if the SSI during the training contains only positive schemas, the model will tend to generate corresponding answers for every label within the SSI during inference.
Therefore, to make the model explicitly reject generating outputs for negative schemas, it is necessary to incorporate negative schemas into the SSI.

In this paper, the schema sequence included in the instructions follows the concept of SSI. However, we observe that existing research ~\citep{DBLP:journals/corr/abs-2304-08085,DBLP:journals/corr/abs-2312-15548} tends to adopt a rather crude schema processing strategy when constructing instructions, meaning that all schemas within a predefined label set are used to build the instructions. This approach potentially entails two significant issues:
1) \textbf{Inconsistency in the number of schema queries within instruction between training and evaluation}.
For example, the model's performance will decrease if it is trained on about 20 schema queries but tested with either 10 or 30, even if the training and evaluation schemas are similar in content.
2) \textbf{Inadequate differentiation among schemas in the instructions}. 
For example, semantically similar schemas like ``layoffs'', ``depart'' and ``dismissals'', may present co-occurrence ambiguities that could confuse the LLMs. Such schemas should co-occur more frequently within the instruction.
Therefore, we introduce: 1) Hard Negative Schema Construction; and 2) Batched Instruction Generation.
Detailed information can be found in Figure \ref{fig:dataconstruct} and Algorithm \ref{alg:alg1}.

\paragraph{Hard Negative Schema Construction.} As illustrated in Figure~\ref{fig:dataconstruct}, assume that dataset $\mathcal{D}$ possesses a predefined label set $L$. 
For a given text $S$, the schemas present in its annotation constitute the positive schema set $Pos\_L$, while others form the negative schema set $Neg\_L$.
In our analysis, we discover that the primary cause of model mistakes stems from the semantic ambiguity of the schema. 
In traditional approaches, the $Neg\_L$ is simply defined as $L - Pos\_L$. 
However, they overlook a critical aspect: it is important to pay special attention to negative schemas that are semantically similar to positive schemas. 
Inspired by the theory of contrastive learning, we construct a hard negative schema dictionary $\mathcal{K}$, where each key represents a unique schema and the associated value is a collection of schemas that are semantically similar to the key schema.
Based on this, we define the hard negative schema set as $Hard\_L = \mathcal{K}[Pos\_L]$, and the other negative schema set as $Other\_L = L - Pos\_L - Hard\_L$.
The final $Neg\_L$ is constituted by $Hard\_L$ and a small subset of $Other\_L$.
Through this strategy, we not only present semantically similar schemas more frequently within the instruction but also reduce the number of training instances without sacrificing model performance.

\paragraph{Batched Instruction Generation.} Subsequently, we obtain the final schema set $L' = Pos\_L + Neg\_L$.
We employ a batched instruction generation method, dynamically limiting the number of schemas inquired in each instruction to the number of $split\_num$, which ranges between 4 and 6. 
Therefore, $L'$ will be divided into $|L'|/split\_num$ batches for querying, with each batch querying $split\_num$ schemas.
Consequently, even if the number of schemas inquired during the evaluation phase differs from that of training, the batched mechanism allows us to distribute the inquiries across $split\_num$ schemas, thereby mitigating the decline in generalization performance.

\begin{figure}[!ht]
\begin{center}
\resizebox{0.48\textwidth}{!}{
\includegraphics{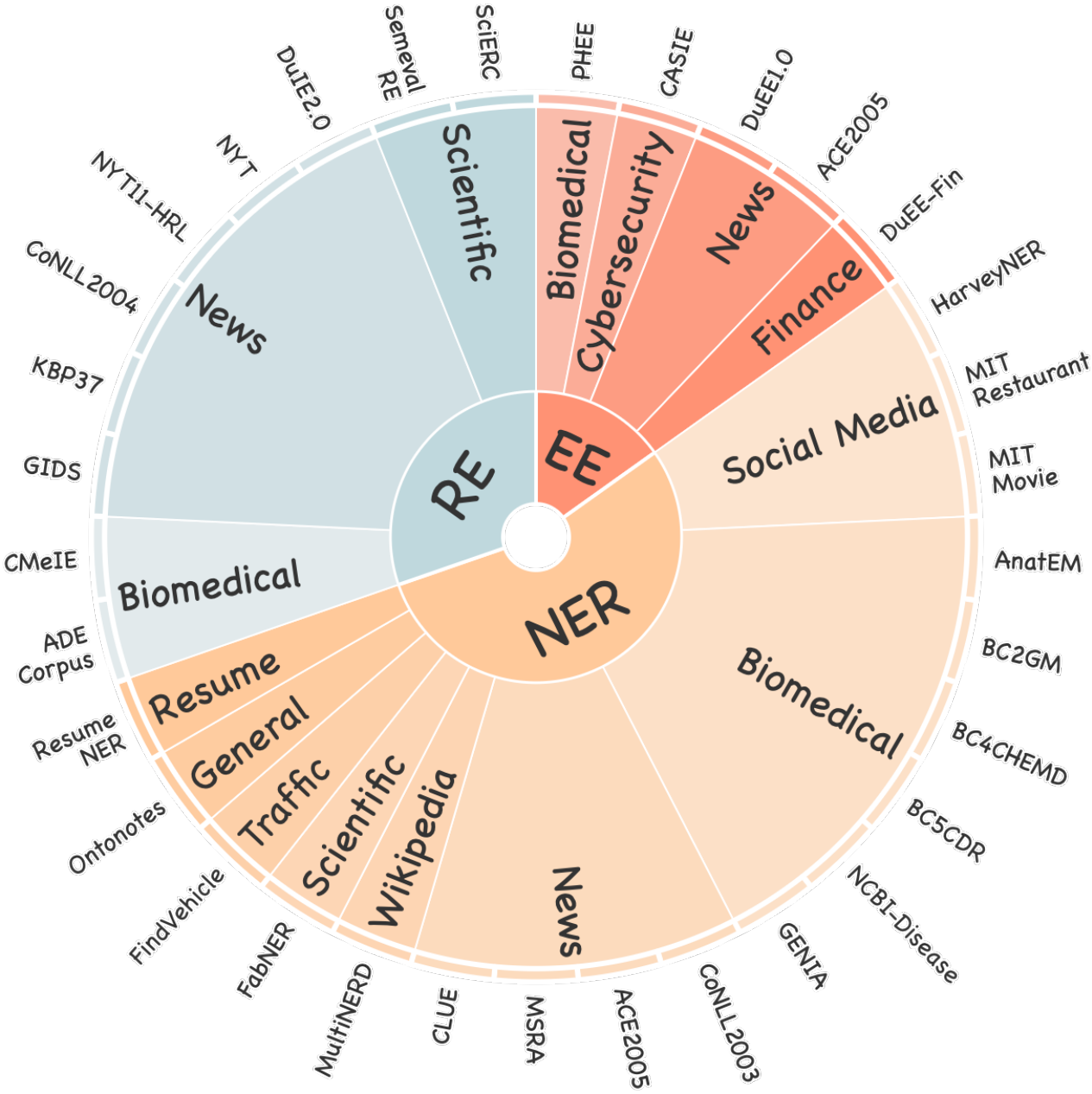}
}
\caption{Distribution of different tasks, domains, and source datasets within the {\ours}.}
\label{fig:statistic}
\end{center}
\end{figure}

\subsection{Data Statistics}
Based on the aforementioned methods, we obtain the {\ours} dataset, which includes roughly 2 million instruction entries and approximately 0.32B tokens (utilizing the Baichuan2 tokenizer).
Figure~\ref{fig:statistic} displays the distribution of domains and source datasets within the {\ours}, including 33 datasets spanning multiple domains such as general, news, finance, and biomedical.
Additionally, Table~\ref{tab:format} provides examples of instructions and outputs for 3 different tasks within the {\ours}.

%% file: experiments.tex
\input{tabs/zero_en}

\section{Experiments}
\label{Experiments}
Based on {\ours}, we fine-tune several latest prominent models, then compare their zero-shot generalization capabilities against a range of baseline models.
Results of the full supervision evaluation and training details are described in Appendix~\ref{apd:experiments}.

\subsection{Experimental Settings}
\textbf{Evaluation Metrics}: We employ span-based Micro-F1 as the metric for measuring model performance. \textbf{Baselines}: We select a range of strong models for comparative analysis, which include UIE, Baichuan2-13B-Chat, GPT-4, InstructUIE, YAYI-UIE, and so on. \textbf{Zero-shot Benchmark}: 
We collect 13 datasets that are \textbf{not present in the training set}. 
\textbf{OneKE}: Additionally, we perform full-parameter fine-tuning of the alpaca2-chinese-13b model utilizing {\ours} and other proprietary information extraction datasets.

\subsection{Main Results}
In Tables~\ref{tab:zero-en} and ~\ref{tab:zero-zh}, we report the zero-shot performance across three tasks and two languages.
Overall, after training with the {\ours}, the models achieve better results in the majority of tasks. 
We believe the success is due to the hard negative schema construction and batched instruction generation strategy, which can mitigate the train-eval mismatch and semantic ambiguity for the diverse schema.
We also observe that {\ours}-models are slightly behind GPT-4 in English NER. 
We hypothesize that the marginal gap may be attributed to GPT-4's exposure to a vast corpus of similar data during its training. 
Moreover, it is essential to note that InstructUIE focuses on English data while {\ours} incorporates both English and Chinese data. 
This disparity in data may influence the capability of the model in English, potentially reducing the performance. 
Additionally, OneKE achieves the best results in nearly all zero-shot evaluation tasks. 
We attribute this success to the enhancements brought by full parameter fine-tuning.

\subsection{Analysis}
\begin{figure}[H]
\begin{center}
\resizebox{0.48\textwidth}{!}{
\includegraphics{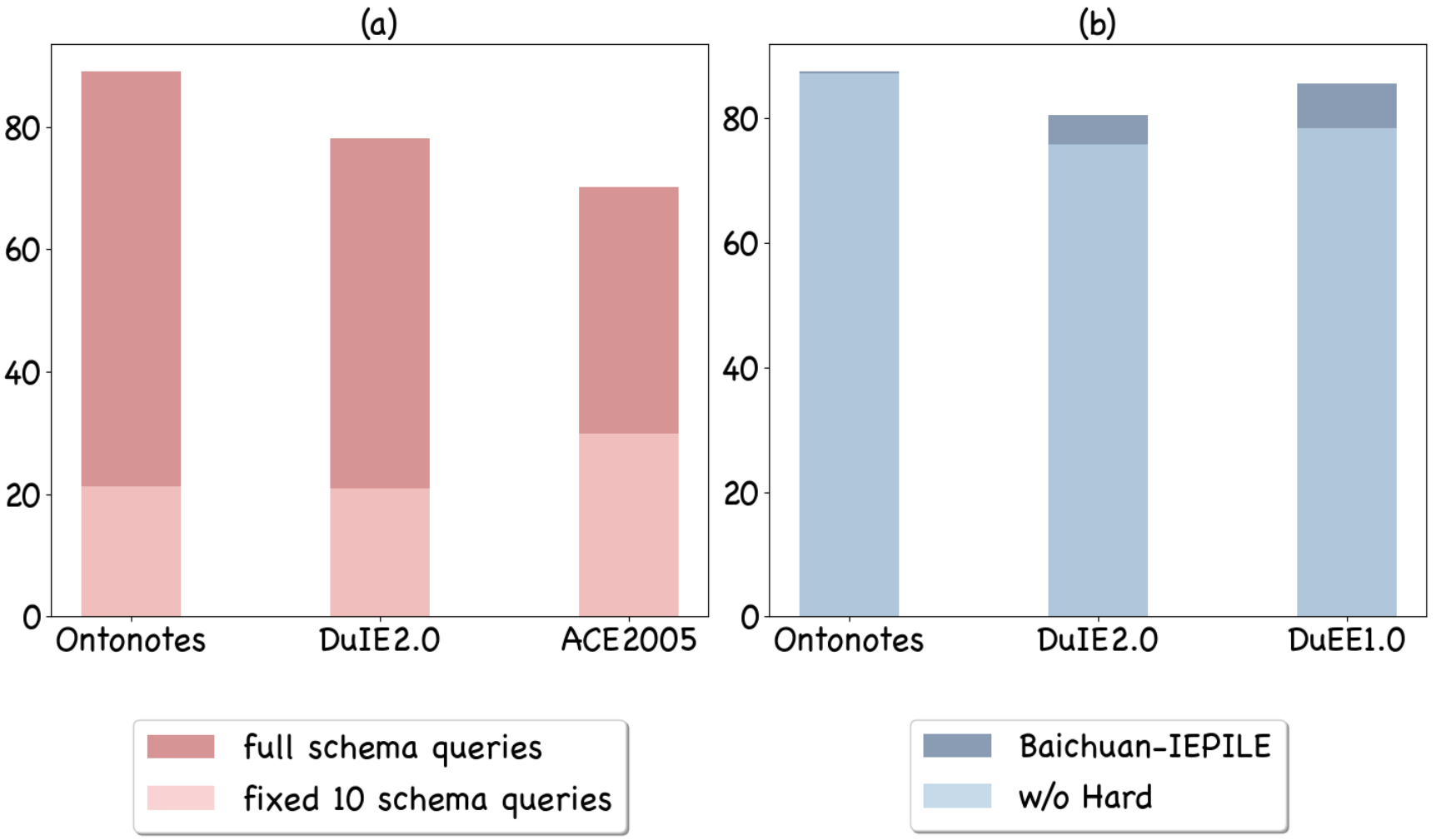}
}
\caption{(a) When there is an inconsistency in the number of schema inquiries during the training and evaluation, the performance of the model significantly decreases. (b) The impact of removing the hard negative schema dictionary on the performance of the model.}
\label{fig:case}
\end{center}
\end{figure}

\paragraph{Inconsistency in the Number of Schema Queries Hurt Generalization.} 
We investigate the impact on model performance when different numbers of schema queries are used during the training and evaluation. 
We train the Baichuan2 using full-schema instructions on 3 datasets: Ontonotes (18 schemas), DuIE2.0 (49 schemas), and ACE2005 (33 schemas). 
For the evaluation, we test the model using two strategies: one with the full set of schema queries and another with a fixed set of 10 schema queries. 
The results depicted in Figure~\ref{fig:case} (a) indicate that the mismatch in the number of schema queries during evaluation significantly reduces the model's performance. 
Further analysis of the model's outputs reveals that the model always tends to generate outputs for each inquiry. 
We hypothesize that the number of schema queries is one of the key factors affecting the generalization ability. 
The model needs to first adapt to the number of schema inquiries that are rare during the training and then adapt to the unseen schema.

\input{tabs/zero_zh}

\paragraph{Inadequate Differentiation Among Schemas Lead to Semantic Similar Confusion.} 
We also evaluate the impact of removing the ``Hard Negative Schema Dictionary'' on the performance of Baichuan2-{\ours}, with particular attention to schemas that are hard to differentiate.
According to the results in Figure~\ref{fig:case} (b), we notice that the hard negative schema dictionary plays a relatively limited role in the NER task, which may be due to the clear boundaries inherent to entity recognition. 
However, the utilization of the hard negative schema dictionary notably enhances model performance in the DuIE2.0 and DuEE1.0 datasets. 
We observe that semantically similar and easily confused schemas frequently appeared in the model's outputs, such as predicting ``dismissal'' and ``resignation'' in the event of ``layoff''. 
Therefore, processing instructions that are semantically prone to confusion poses significant challenges, and the hard negative schema dictionary plays a crucial role in bolstering model robustness and improving the accuracy of predictions.

%% file: tabs/zero_en.tex
\begin{table*}[t]
    \centering
    \setlength{\tabcolsep}{2.5pt}
    \begin{tabular}{l|c|cc|c|ccc|c} 
    \toprule
        \multirow{2}*{Method} & \multicolumn{1}{c|}{NER} & \multicolumn{3}{c|}{RE} & \multicolumn{4}{c}{EE}  \\
        \cline{2-9}
         ~ & CrossNER & FewRel & Wiki-ZSL & Avg & WikiEvents & RAMS & CrudeOil News & Avg \\
        \hline
        LLaMA2 & 34.82  & 6.53 & 9.43 & 7.98 & 0.00 & 0.00 & 0.00 & 0.00  \\
        Baichuan2 & 38.93 & 5.94 & 4.15 & 5.05 & 0.00 & 0.00 & 0.00 & 0.00 \\
        Qwen1.5 & 50.13 & 7.82 & 6.94 & 7.38 & 0.00 & 0.00 & 0.00 & 0.00 \\
        Mistral & 42.83 & 6.84 & 5.10 & 5.97 & 0.00 & 0.00 & 0.00 & 0.00 \\
        ChatGPT & 58.37  & 9.96 & 13.14 & 11.55  & 2.95 & 8.35 & 1.41 & 4.24  \\
        GPT-4 & \CB 58.49 & 22.43 & 23.76 & 23.10 & 5.24 & 10.14 & 26.13 & 13.84 \\
        \hline 
        UIE & 38.37 & - & - & -  & 5.12 & 9.25 & 6.45 & 6.94  \\
        InstructUIE & 49.36 & 39.55 & 35.20 & 37.38 & 11.64 & \CC \textbf{24.27} & 23.26 & 19.72 \\ 
        YAYI-UIE  & 50.39  & 36.09 & \CB 41.07 & 38.58 & 10.97 & 18.87 & 12.45 & 14.10   \\
        \hline
        Baichuan2-{\ours}  & 55.55  & \CC \textbf{41.28} & 37.61 & 39.45 & 9.12 & 20.19 & 36.61 & 21.97   \\
        LLaMA2-{\ours} &  56.50  & 37.14 & 36.18  & 36.66 & \CC \textbf{13.93} & \CB 23.62 & 33.87 & \CB 23.81  \\ 
        Qwen1.5-{\ours} &  57.90  & \CB 40.92 & 38.49  & \CB 39.71 & 11.38 & 21.26 & 30.69 & 21.11  \\ 
        LLaMA3-{\ours} & 56.11 & 35.58 & 37.18 & 36.38 & 9.71 & 20.27 & \CC \textbf{39.88} & 23.29 \\
        OneKE & \CC \textbf{60.91} & 39.19 & \CC \textbf{42.18} & \CC \textbf{40.68} & \CB 12.43 & 22.58 & \CB 38.49 & \CC \textbf{24.50} \\
        \bottomrule
    \end{tabular}
    \caption{Zero-shot performance on English datasets. UIE necessitates predefined entity types; given that such information is not provided by the FewRel and Wiki-ZSL datasets, we are unable to evaluate UIE's performance on these datasets. For the task of event extraction, we only present the results of event detection in the main text.}
    \label{tab:zero-en}
\end{table*}

%% file: tabs/zero_zh.tex
\begin{table*}[t]
    \centering
    \setlength{\tabcolsep}{2pt}
    \begin{tabular}{l|cc|c|ccc|c|cc|c} 
    \toprule
        \multirow{2}*{Method} & \multicolumn{3}{c|}{NER} & \multicolumn{4}{c|}{RE} & \multicolumn{3}{c}{EE} \\
        \cline{2-11}
         ~ & Boson & Weibo & Avg & SKE2020 & COAE2016 & IPRE  & Avg & FewFC  & CCF Law & Avg \\
        \hline
        LLaMA2 & 8.19 & 2.43 & 5.31 & 0.50 & 3.11 & 0.31  & 1.31 & 0.23 & 0.08 & 0.16 \\
        Baichuan2 & 27.39 & 7.62 & 17.51 & 7.23 & 11.65 & 1.45 & 6.78 & 11.82 & 2.73 & 7.28 \\
        Qwen1.5 & 26.49 & 25.34 & 25.92 & 7.69 & 11.97 & 2.16 & 7.27 & 11.47 & 3.25 & 7.36 \\
        Mistral & 29.13 & 10.02 & 19.58 & 6.84 & 5.24 & 0.82 & 4.30 & 4.69 & 0.23 & 2.46 \\
        ChatGPT & 38.53 & 29.30 & 33.92 & 24.47 & 19.31 & 6.73 & 16.84 & 16.15 & 0.00 & 8.08 \\
        GPT-4 & 48.15 & 29.80 & 38.98 & 56.77 & 41.15 & 18.15 & 38.69 & 74.25 & 42.12 & 58.19 \\
        \hline 
        YAYI-UIE  & 49.25 & 36.46  & 42.86 & 70.80 & 19.97 & 22.97 & 37.91 & 81.28 & 12.87 & 47.08 \\
        \hline
        Baichuan2-{\ours}  & 55.77 & \CC \textbf{38.03}  & 46.90 & 72.50 & 47.43 & 29.76  & 49.90 & \CC \textbf{83.59}  & \CC \textbf{63.53} & \CC \textbf{73.56} \\
        LLaMA2-{\ours}  & 54.45 & 34.97  & 44.71  & 72.18  & 46.70  & 28.55 & 49.14  & 70.10 & 59.90 & 65.00 \\
        Qwen1.5-{\ours}  & \CB 63.08 & \CB 37.50  & \CB 50.29  & 72.29  & \CC \textbf{50.70}  & \CB 30.55 & \CB 51.18  & 78.77 & 61.43 & 70.10 \\
        LLaMA3-{\ours} & 61.88 & 37.43 & 49.66 & \CB 73.67 & 48.12 & \CC \textbf{31.29} & 51.03 & \CB 81.52 & 59.92 & 70.72 \\
        OneKE & \CC \textbf{72.61} & 35.06 & \CC \textbf{53.84} & \CC \textbf{74.15} & \CB 49.83 & 29.95 & \CC \textbf{51.31} & 80.11 & \CB 62.19 & \CB 71.15 \\
        \bottomrule
    \end{tabular}
    \caption{Zero-shot performance on Chinese datasets. Since UIE and InstructUIE do not train with Chinese data, we do not report performance of these two models on Chinese datasets.}
    \label{tab:zero-zh}
\end{table*}

%% file: appendix.tex
\appendix

\newpage

\input{related_work}

\section{Construction Details of {\ours}}
\label{apd:iepile}

\subsection{Data Collection and Clean}
\label{apd:data-collection-clean}

\paragraph{Data Collection} To comprehensively cover various domains and meet the practical demands of information extraction (IE), we collect IE datasets from multiple sources. 
{\ours} dataset mainly involves bilingual data (Chinese and English) and three IE tasks: Named Entity Recognition (NER), Relation Extraction (RE), and Event Extraction (EE). 
The English part mainly comes from the benchmark dataset IEINSTRUCTIONS ~\citep{DBLP:journals/corr/abs-2304-08085}, while the Chinese data is similar to the Chinese datasets mentioned in the YAYI-UIE ~\citep{DBLP:journals/corr/abs-2312-15548}. 
It should be noted that our Chinese dataset collection is conducted concurrently with the aforementioned research. 

Specifically, the NER datasets include fifteen English datasets such as ACE2005 ~\citep{ace2005-annotation}, AnatEM ~\citep{DBLP:journals/bioinformatics/PyysaloA14}, BC2GM ~\citep{DBLP:conf/icpr/KocamanT20}, BC4CHEMD ~\citep{DBLP:conf/icpr/KocamanT20}, BC5CDR ~\citep{DBLP:conf/iclr/00120GP23}, CoNLL2003 ~\citep{DBLP:conf/conll/SangM03}, FabNER ~\citep{DBLP:journals/jim/KumarS22}, FindVehicle ~\citep{DBLP:journals/corr/abs-2304-10893}, GENIA-Ent ~\citep{DBLP:conf/ismb/KimOTT03}, HarveyNER ~\citep{DBLP:conf/naacl/ChenXZH22}, MIT Movie ~\citep{DBLP:conf/icassp/LiuPCG13}, MIT Restaurant ~\citep{DBLP:conf/icassp/LiuPCG13}, MultiNERD ~\citep{DBLP:conf/naacl/TedeschiN22}, NCBI-Disease ~\citep{DBLP:journals/jbi/DoganLL14}, Ontonotes ~\citep{DBLP:conf/naacl/PradhanX09}, and three Chinese datasets including MSRA ~\citep{DBLP:conf/acl-sighan/Levow06}, Resume NER ~\citep{DBLP:conf/acl/ZhangY18}, CLUE NER ~\citep{DBLP:journals/corr/abs-2001-04351}. The RE task encompasses eight English datasets including ADE Corpus ~\citep{DBLP:journals/biomedsem/GurulingappaRT12}, CoNLL2004 ~\citep{DBLP:conf/conll/CarrerasM04}, GIDS ~\citep{DBLP:conf/akbc/JatKT17}, KBP37 ~\citep{DBLP:journals/corr/ZhangW15a}, NYT ~\citep{DBLP:conf/pkdd/RiedelYM10}, NYT11-HRL  ~\citep{DBLP:conf/aaai/TakanobuZLH19}, SciERC ~\citep{DBLP:conf/emnlp/LuanHOH18}, Semeval-RE ~\citep{DBLP:conf/semeval/HendrickxKKNSPP10}, and two Chinese datasets, CMeIE ~\citep{DBLP:conf/emnlp/LuanHOH18}, DuIE2.0 ~\citep{DBLP:conf/semeval/HendrickxKKNSPP10}. The EE task covers three English datasets: ACE2005 ~\citep{ace2005-annotation}, CASIE ~\citep{DBLP:conf/aaai/SatyapanichFF20}, PHEE ~\citep{DBLP:conf/emnlp/Sun0PWJGK022}, and two Chinese datasets, DuEE1.0 ~\citep{DBLP:conf/aaai/SatyapanichFF20}, DuEE-fin ~\citep{DBLP:conf/emnlp/Sun0PWJGK022}. 
These datasets span various domains such as general, medical, financial, and more. 
For more detailed statistical information, please refer to Tables~\ref{statistic-ner}, ~\ref{statistic-re} and ~\ref{statistic-ee}.

\paragraph{Data Cleaning} 
During the data cleaning process, we address each dataset individually.
Firstly, we calculate the text overlap within each dataset’s training, validation, and test sets. 
If a text is discovered to have multiple occurrences within the same file accompanied by inconsistent annotations, we exclude all corresponding instances from the dataset.
Secondly, we compare the text overlap between training, validation, and test sets. 
If texts from the test set appear previously in the training or validation sets, we would exclude these instances from the training and validation sets. 
Furthermore, we formulate three heuristic rules to eliminate low-quality and meaningless data:

\textbf{1)} Non-alphabetic characters comprising more than 80\% of the text; 

\textbf{2)} Text length under five characters without any labels;

\textbf{3)} A high prevalence of stopwords such as `the,' `to,' `of,' etc., exceeding 80\%.

\begin{figure}[!ht]
\begin{center}
\resizebox{0.48\textwidth}{!}{
\includegraphics{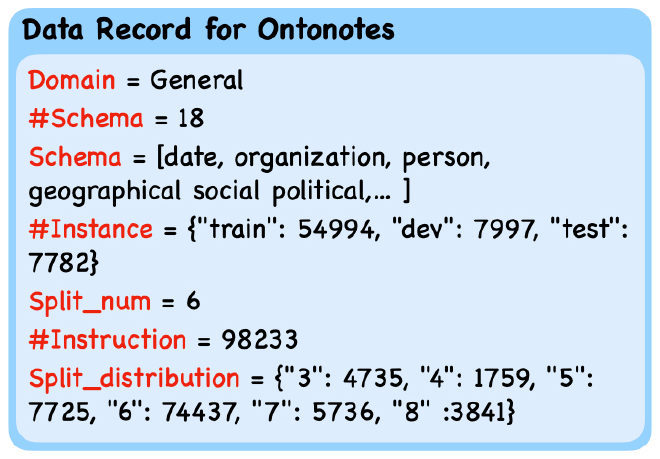}
}
\caption{An exemplar of data records for OntoNotes: the domain, the number and details of schemas, the total volume of data, the $split\_num$, the number of instructions produced using our method, along with the distribution of split count within the interval [($split\_num$ / 2), ($split\_num$ + $split\_num$ / 2)].}
\label{fig:record}
\end{center}
\end{figure}

We believe that the aforementioned cleaning measures will positively affect model training and enhance its performance. 
Moreover, our efforts unify data formats across various tasks and conduct a thorough audit of each dataset, creating detailed \textbf{data records} that include the volume of data, domains, schemas, and other information. 
Figure~\ref{fig:record} is an example of a data record for Ontonotes.

\subsection{Schema-Based Instruction Generation}
\label{apd:instruction-generation}

\paragraph{Hard Negative Schema Construction.}As illustrated in Figure~\ref{fig:dataconstruct}, assume that dataset $\mathcal{D}$ possesses a predefined label set $L$. 
For a given text $S$, the schemas present in its annotation constitute the positive schema set $Pos\_L$, while others form the negative schema set $Neg\_L$.
Inspired by the theory of contrastive learning, we construct a hard negative schema dictionary $\mathcal{K}$, where each key represents a unique schema and the associated value is a collection of schemas that are semantically similar to the key schema.
Consequently, the set of hard negative schema, $Hard\_L$, is defined as $\mathcal{K}[Pos\_L]$. 
However, if $Neg\_L$ is composed solely of $Hard\_L$, it would lack a sufficient number of negative instances for the model to learn effectively. 
Therefore, we define another set of negative schemas, $Other\_L = L - Hard\_L - Pos\_L$. 
Ultimately, the $Neg\_L$ is composed of $Hard\_L$ and a small number of $Other\_L$ (roughly $split\_num$). 
The rationale behind the development of these hard negatives is two-fold: firstly, to induce a more frequent co-occur of semantically similar schemas within the instructions, and secondly, to reduce the volume of training instances without sacrificing the model's performance.
In the context of a dataset comprising 48 schemas with a given $split\_num$ of 4, traditional mode would dictate the creation of 12 unique instructions per data point. 
However, through the integration of hard negatives, this requisite can be substantially minimized to a mere 3 instructions.

\input{dataconstruct_alg}

\paragraph{Batched Instruction Generation.}
Subsequently, we obtain the final schema set $L' = Pos\_L + Neg\_L$. 
During the instruction generation phase, the role of schemas is critically vital, as it reflects the specific extraction requirements and is dynamically variable.
Traditional practices typically integrate the full schema set into the instruction. 
However, in this study, we employ a batched instruction generation method, dynamically limiting the number of schemas inquired in each instruction to the number of $split\_num$, which ranges between 4 to 6. 
Therefore, $L'$ will be divided into $|L'|/split\_num$ batches for querying, with each batch querying $split\_num$ schemas.
Consequently, even if the number of schemas inquired during the evaluation phase differs from that of training, the batched mechanism allows us to distribute the inquiries across $split\_num$ schemas, thereby mitigating the decline in generalization performance.

\paragraph{Selection of split\_num.}
In the determination of the optimal range for $split\_num$, our methodology integrates empirical results with an in-depth analysis of dataset characteristics.
For a dataset containing N different labels, the theoretical value of $split\_num$ should fall within the interval [1, N]. 
Addressing datasets with heterogeneous label counts, our objective is to identify a $split\_num$ value that offers broad applicability across numerous datasets, thus ensuring this value serves as a common divisor for the majority of dataset label counts.
For instance, for Named Entity Recognition datasets, we set $split\_num$ to 6; for Relation Extraction and Event Extraction datasets, we establish $split\_num$ at 4.
We also observe that when $split\_num$ is 1, the ratio of positive to negative samples significantly impacts model performance, and the corresponding number of training samples becomes vast, affecting efficiency adversely. 
More crucially, we believe that enumerating multiple schemas in instructions aids the model in more effectively learning to distinguish and identify various schemas, thereby enhancing model performance.

Furthermore, to enhance model robustness and its clear understanding of the dynamically changing schema sequences in instructions, we set the actual number of schema splits within a dynamic range of [$split\_num$ // 2, $split\_num$ + $split\_num$ // 2]. 
Specifically, if the number of schemas in the last batch is less than half of $split\_num$, it is merged with the previous batch; otherwise, it stands as an independent batch.

\paragraph{Instruction Format}
The instruction format of {\ours} adopts a structure akin to JSON strings, essentially constituting a dictionary-type string. 
This structure is comprised of three main components: (1) ``instruction'', which is the task description outlining the objective of the instruction's execution; (2) ``schema'', a list of labels that need to be extracted; (3) ``input'', the source text from which information is to be extracted.
Examples of instructions corresponding to various tasks can be found in Table~\ref{tab:format}.

\subsection{Data Statistics}
Based on the aforementioned methodologies, we construct a high-quality information extraction instruction dataset known as {\ours}. 
This dataset contains approximately two million instances and approximately 0.32B tokens.
Each instance of {\ours} comprises two fields: ``instruction'' and ``output'', which are formatted for direct use in the instruction tuning.

\section{Experiments}
\label{apd:experiments}

\subsection{Experimental Settings}

\paragraph{Evaluation Metrics}
We employ span-based Micro-F1 as the primary metric for measuring model performance. 
For the NER task, the model is required to accurately identify the boundaries of entities and their corresponding types. 
For the RE task, the model must precisely determine the subject and object entities within a relation, as well as the type of relation between them. 
As for the EE task, we match the event triggers, denoted as \textbf{Trigger}, and the arguments, referred to as \textbf{Argument}, independently.

\paragraph{Baselines}
To assess the zero-shot generalization capabilities, we select a range of strong models for comparative analysis:
\begin{itemize}
  \item UIE ~\citep{DBLP:conf/acl/0001LDXLHSW22}: is a unified text-to-structure generation framework that can model various information extraction (IE) tasks generically. 
  \item LLaMA2 ~\citep{DBLP:journals/corr/abs-2307-09288}: is a series of LLMs ranging from 7 billion to 70 billion parameters.
  \item Baichuan2 ~\citep{DBLP:journals/corr/abs-2309-10305}: is a collection of multilingual LLMs containing 7 billion and 13 billion parameters.
  \item Qwen1.5 ~\citep{qwen}: is a comprehensive language model series that encompasses distinct models with varying parameter counts.
  \item Mistral ~\citep{DBLP:journals/corr/abs-2310-06825}: is a 7-billion-parameter LLM.
  \item ChatGPT ~\citep{ouyang2022training}: also known as GPT-3.5-turbo, represents the most advanced artificial intelligence language model with chat optimization capabilities to date.
  \item GPT-4 ~\citep{DBLP:journals/corr/abs-2303-08774}: Known as the most powerful closed-source chat model to date.
  \item LLaMA3 \footnote{\url{https://ai.meta.com/blog/meta-llama-3/}.}: The latest release in the LLaMA model series, achieving significant improvements across various benchmarks.
  \item InstructUIE ~\citep{DBLP:journals/corr/abs-2304-08085}:  a unified IE framework based on multi-task instruction tuning.
  \item YAYI-UIE ~\citep{DBLP:journals/corr/abs-2312-15548}: is an end-to-end, chat-enhanced, universal information extraction framework that supports both Chinese and English, fine-tuned with instructional prompts for generalized information.
\end{itemize}

\subsection{OneKE}
We leverage {\ours}, InstructIE \citep{DBLP:journals/corr/abs-2305-11527}, CMRC \citep{DBLP:conf/emnlp/CuiLCXCMWH19}, along with certain proprietary business information extraction datasets from Ant Group, to compile a comprehensive training dataset consisting of 2.5 million instances.
Subsequently, we undertake full-parameter fine-tuning of the alpaca2-chinese-13b\footnote{\url{https://huggingface.co/hfl/chinese-alpaca-2-13b}.} model on this training dataset, resulting in the refined model named OneKE.

\input{tabs/zero_ee}

\paragraph{Zero-shot Dataset}
To ensure the validity of the zero-shot evaluation and prevent result bias due to data similarity, we select datasets primarily derived from news and biomedical fields as our training sets.
This selection is intended to train the model's capability for instruction following and schema-based extraction. 
For the evaluation data, we adopt the 13 cross-domain datasets recommended in IEINSTRUCTIONS and YAYI-UIE, which include: for Named Entity Recognition (NER) tasks, we use the CrossNER ~\citep{DBLP:conf/aaai/Liu0YDJCMF21}, Weibo NER ~\citep{DBLP:conf/emnlp/PengD15}, and Boson\footnote{\url{https://github.com/InsaneLife/}\label{fn:boson}}; in Relation Extraction (RE) tasks, we choose FewRel ~\citep{DBLP:conf/emnlp/HanZYWYLS18}, Wiki-ZSL ~\citep{DBLP:conf/naacl/ChenL21}, COAE2016\footnote{\url{https://github.com/Sewens/COAE2016}\label{fn:coae2016}}, IPRE ~\citep{wang2019ipre}, and SKE2020\footnote{\url{https://aistudio.baidu.com/datasetdetail/177191}\label{fn:SKE2020}}; and for Event Extraction (EE), we include RAMS ~\citep{ebner-etal-2020-multi}, WikiEvents ~\citep{DBLP:conf/naacl/LiJH21}, CrudeOilNews ~\citep{DBLP:conf/lrec/LeeSSS22}, FewFC ~\citep{DBLP:conf/aaai/Zhou0ZWXL21}, and CCF law \footnote{\url{https://aistudio.baidu.com/projectdetail/4201483}\label{fn:ccflaw}}. 
These datasets cover a wide range of fields including literature, music, law, and oil news.
It is noteworthy that these evaluation data sets are not used during the training, ensuring that our evaluation accurately reflects the model's generalization and adaptation capabilities for unseen domains and unseen schema data in zero-shot information extraction.

\subsection{Zero-shot performance on Event Extraction}
As illustrated in Table ~\ref{tab:zero-ee}, the model trained with {\ours} exhibits outstanding performance in zero-shot event extraction (EE) tasks, surpassing other baselines.
Notably, in the Chinese EE task, the LLaMA2-{\ours} model's performance is slightly inferior to YAYI-UIE's, revealing LLaMA2's limitations in processing Chinese data. 
However, in the English EE task, LLaMA2-{\ours}'s performance is significantly superior to that of similar models. 
This contrast highlights the potential influence of language type on model performance.

\begin{table}[ht!]
\centering
\begin{tabular}{l|l}
\toprule
Hyperparameter              & Value         \\ \hline
Number of Epochs            & 5             \\
Learning Rate               & 5e-5          \\
Batch Size                  & 20            \\
Accumulate                  & 4             \\
Optimizer                   & Adam          \\
Lora\_r                     & 64            \\
Lora\_alpha                 & 64            \\  
Lora\_dropout               & 0.05          \\
\bottomrule
\end{tabular}
\caption{Training Hyperparameters}
\label{tab:hyperparameters}
\end{table}

\subsection{Hyper-parameter}
In our research, we select four pre-trained models, Baichuan2-13B-Chat and LLaMA2-13B-Chat, Qwen1.5-14B-Chat, and LLaMA3-8B-Instruct, as the base models for our study. 
Specifically, we employ the LoRA ~\citep{DBLP:conf/iclr/HuSWALWWC22} technique and utilize 8 NVIDIA A800 GPUs to perform instruction tuning on our {\ours} dataset. 
Detailed configurations of the hyperparameters during the fine-tuning process are presented in Table~\ref{tab:hyperparameters}.

\begin{table}[t!]
\centering
\begin{tabular}{l|cc}
\toprule
\textbf{Dataset}     & \textbf{Supervised}  & \textbf{Zero-shot} \\ \hline
ACE2004                & 84.28      & 77.01      \\ 
People Daily           & 98.34      & 95.29      \\ 
\bottomrule
\end{tabular}
\caption{The results of individual LoRA fine-tuning on ACE2004 and People Daily datasets for Baichuan2-13B-Chat, compared with the zero-shot generalization results of Baichuan2-{\ours} on these two datasets.}
\label{tab:similar_dataset}
\end{table}

\subsection{Supervision Results}
Due to limited computational resources, I report only the supervised results for the Baichuan2-{\ours}, LLaMA2-{\ours}, and OneKE models.
Tables ~\ref{supervised-ner}, ~\ref{supervised-re}, and ~\ref{supervised-ee} present our experimental results under a supervised learning setting on the training dataset. 
Specifically, it can be observed that after training on the {\ours}, the model excels in Named Entity Recognition (NER), Relation Extraction (RE), and Event Detection (ED), ranking top 2 across these tasks. 
The model's performance is only slightly behind other baselines in the Event Argument Extraction. 
Additionally, we record the model's performance in Chinese NER, RE, and EE tasks, where it demonstrates robust results. 
In a comprehensive assessment, the {\ours}-trained model showcases performance on par with other models in instruction-based information extraction (IE) tasks and significantly improves performance in zero-shot IE tasks compared to other models. 
This indicates the significant application prospects and potential of {\ours} in the current field of IE.

\subsection{Impact of Potential Dataset Bias on Model Performance and Generalization}
During the research, we identify that potential biases introduced by the datasets used can affect the model's performance and generalization capability. 
Firstly, biases in the definition of schemas within the datasets have a negative impact on model performance ~\citep{huang2024onemodelfitsall}. 
In the early stages of training, we observe instability in results due to mutual interference among multiple datasets that contain the same schemas but with differing definitions. 
For instance, despite wikiann, wikineural, polyglot-NER, and CoNLL2003 all containing common schemas such as people and organization, they each possess distinct scheme definitions. 
Consequently, in the later stages, only CoNLL2003 is retained. 
Secondly, the model demonstrates good generalization when dealing with datasets having schemas similar to those in the training set. 
As shown in Table \ref{tab:similar_dataset}, despite not being included in the training corpus, the People Daily and ACE2004 NER datasets share similar schemas with the MASR and ACE2005 NER dataset in the training set, and the Baichuan2-{\ours} model is still capable of handling them proficiently.
Lastly, the use of common, coarse-grained labels (such as ``person'' and ``organization'') within the {\ours} lead the model, after training, to favor these coarse categories over fine-grained ones (such as ``scientist'' and ``company'') when predicting instructions that included both levels of granularity.

\input{tabs/supervised_ner}
\input{tabs/supervised_re}

\input{tabs/supervised_ee}

\input{tabs/statistic_ner}

\input{tabs/statistic_re}
\input{tabs/statistic_ee}

\input{tabs/instructions}

%% file: related_work.tex
\section{Related Work}

\subsection{Information Extraction Datasets}
Large-scale pre-trained corpora are crucial for the effectiveness of LLMs, providing a wealth of knowledge and a foundation for language comprehension. 
At the same time, the annotated data for information extraction (IE) also holds its importance.
Although the field of IE has accumulated a considerable amount of annotated data~\citep{ace2005-annotation,DBLP:conf/pkdd/RiedelYM10,DBLP:conf/conll/SangM03,DBLP:conf/emnlp/LuanHOH18,DBLP:journals/corr/abs-2305-11527}, these datasets are often limited in size, scattered in distribution, and lack standardization in schema.
Faced with these limitations, there is an urgent need for generating instruction data through unified and automated methods to bridge the gap presented by the current absence of centralized, large-scale IE instruction datasets. 
In this paper, we concentrate on instruction-based IE scenarios. 
We develop a comprehensive, schema-rich instruction dataset for IE by collecting and cleaning existing IE datasets, called {\ours}. 
{\ours} is designed to enhance the adaptability and processing capabilities of LLMs for different IE tasks, simultaneously strengthening their generalization skills to extract from new domains and schemas.

\subsection{Information Extraction Models}
Recently, LLMs~\citep{brown2020language,ouyang2022training,touvron2023llama,DBLP:journals/corr/abs-2307-09288} demonstrate their exceptional versatility and generalization capabilities across a variety of downstream tasks ~\citep{DBLP:conf/acl/VilarFCLRF23,DBLP:conf/aistats/HegselmannBLA0S23}. 
Particularly in the domain of IE, these models have the potential to tackle many challenges previously encountered in research ~\citep{zheng-etal-2017-joint,li-etal-2020-unified,paolini2021structured,lu-etal-2022-unified,DBLP:conf/aaai/Lou0DJLH0023,DBLP:conf/www/ChenZXDYTHSC22,chen2024sequence}, such as adaptability issues when dealing with unseen labels. 
Some studies ~\citep{DBLP:journals/corr/abs-2302-10205,DBLP:journals/corr/abs-2304-10428,DBLP:journals/corr/abs-2311-08921} make significant performance gains in low-resource settings by designing prompt-based frameworks and leveraging models like ChatGPT for in-context learning.
Moreover, research efforts such as InstructUIE ~\citep{DBLP:journals/corr/abs-2304-08085}, PIVOINE ~\citep{DBLP:journals/corr/abs-2305-14898}, and YAYI-UIE ~\citep{DBLP:journals/corr/abs-2312-15548}, which employ instruction-tuning of open-source LLMs, also achieve notable successes on IE. 
Additional research explore areas such as prompt learning ~\citep{DBLP:conf/emnlp/ZhangLLGWHL23}, guidelines ~\citep{DBLP:journals/corr/abs-2310-03668} and synthetic dataset ~\citep{DBLP:conf/emnlp/AmalvyLD23}. 
Despite these advancements, current models fine-tuned with instruction data face a major challenge: the coarse schema handling strategies in constructing instructions could potentially impair the models' capacity for generalization.

%% file: dataconstruct_alg.tex
\begin{algorithm}[t]
\caption{Schema-Based Instruction Generation}
\begin{algorithmic}
\REQUIRE Text $S$, Predefined label set $L$, Positive schema set $Pos\_L$, Number of schemas to split $split\_num$
\ENSURE Set of $Instructions$

\textbf{Step 1: Initialize Hard Negative Schema Dictionary $\mathcal{K}$}
\FORALL{schema in $L$}
    \STATE $\mathcal{K}[schema] \leftarrow \text{SEMANTIC-SIMILAR}(schema, L)$
\ENDFOR

\STATE \textbf{Step 2: Obtain Hard Negative Schemas}
\STATE $Hard\_L \leftarrow \emptyset$  
\FORALL{schema in $Pos\_L$}
    \STATE $Hard\_L \leftarrow Hard\_L \cup \mathcal{K}[schema]$
\ENDFOR

\STATE $Other\_L \leftarrow L - Pos\_L - Hard\_L$
\STATE $Other\_L \leftarrow \text{RANDOM-SELECT}(Other\_L, split\_num)$
\STATE $Neg\_L \leftarrow Hard\_L \cup Other\_L$
\STATE $L' \leftarrow Neg\_L \cup Pos\_L$
\STATE Shuffle $L'$ to obtain a randomized sequence

\STATE \textbf{Step 3: Batched Instruction Generation}
\STATE $Instructions \leftarrow []$
\STATE $num\_batches \leftarrow \lceil \frac{|L'|}{split\_num} \rceil$
\FOR{$i \leftarrow 1$ \TO $num\_batches$}
    \STATE $Batch \leftarrow \text{SEQUENTIAL-SELECT}(L', split\_num, i)$
    \STATE $Instructions \leftarrow Instructions \cup \text{GENERATE-INSTRUCTION}(Batch)$
\ENDFOR
\end{algorithmic}
\label{alg:alg1}
\end{algorithm}

%% file: tabs/zero_ee.tex
\begin{table*}[t] 
    \centering
    \setlength{\tabcolsep}{3pt}
    \begin{tabular}{c|c|ccc|c|cc|c} 
    \toprule
        \multirow{2}*{~} &\multirow{2}*{Method} & \multicolumn{4}{c|}{EN} & \multicolumn{3}{c}{CH}  \\
        \cline{3-9}
        ~ & ~  & WikiEvents & RAMS & \begin{tabular}[c]{@{}c@{}}CrudeOil\\News\end{tabular} & Avg & FewFC & \begin{tabular}[c]{@{}c@{}}CCF\\Law\end{tabular} & Avg \\
        \hline
        \multirow{4}*{Trigger}  
        & LLaMA2 & 0.00 & 0.00 & 0.00 & 0.00 & 0.23 & 0.08 & 0.16 \\
        & Baichuan2 & 0.00 & 0.00 & 0.00 & 0.00 & 11.82 & 2.73 & 7.28 \\
        & Qwen1.5 & 0.00 & 0.00 & 0.00 & 0.00 & 11.47 & 3.25 & 7.36 \\
        & Mistral & 0.00 & 0.00 & 0.00 & 0.00 & 4.69 & 0.23 & 2.46  \\
        & ChatGPT & 2.95 & 8.35 & 1.41 & 4.24 & 16.15 & 0.00 & 8.08 \\
        & GPT4.0 & 5.24 & 10.14 & 26.13 & 13.84 & 74.25 & 42.12 & 58.19 \\
        \hline 
        & UIE  & 5.12 & 9.25 & 6.45 & 6.94 & - & - & - \\
        & InstructUIE & 11.64 & \CC \textbf{24.27} & 23.26 & 19.72 & - & - & - \\
        & YAYI-UIE   &  10.97 & 18.87 & 12.45 & 14.10 & 81.28 & 12.87 & 47.08\\
        \cline{2-9}
        & Baichuan2-{\ours}  & 9.12 & 20.19 & 36.61 & 21.97 & \CC \textbf{83.59}  & \CC \textbf{63.53} & \CC \textbf{73.56} \\
        & LLaMA2-{\ours} & \CC \textbf{13.93} & \CB 23.62 & 33.87 & \CB 23.81 & 70.10 & 59.90 & 65.00 \\
        & Qwen1.5-{\ours} & 11.38 & 21.26 & 30.69 & 21.11 & 78.77 & 61.43 & 70.10 \\
        & LLaMA3-{\ours} & 9.71 & 20.27 & \CC \textbf{39.88} & 23.29 & \CB 81.52 & 59.92 & 70.72 \\
        & OneKE & \CB 12.43 & 22.58 & \CB 38.49 & \CC \textbf{24.50} & 80.11 & \CB 62.19 & \CB 71.15 \\
        \hline
        \multirow{4}*{Argument}   
        & LLaMA2 & 0.00 & 0.00 & 0.00 & 0.00 & 0.00 & 0.06 & 0.03 \\
        & Baichuan2 & 0.79 & 1.81 & 0.48 & 1.03 & 6.91 & 13.04 & 9.98 \\
        & Qwen1.5 & 0.64 & 2.31 & 0.74 & 1.23 & 6.37 & 14.48 & 10.43 \\
        & Mistral & 0.24 & 0.65 & 0.16 & 0.35 & 7.43 & 6.60 & 7.02 \\
        & ChatGPT & 2.07 & 2.21 & 8.60 & 4.29 & 44.40 & 44.57 & 44.49 \\
        & GPT4.0 & 3.35 & 7.35 & 17.25 & 9.32 & 48.05 & 47.49 & 47.77 \\
        \hline 
        & UIE & 1.78 & 2.14 & 8.95 & 4.29 & - & - & - \\
        & InstructUIE & 5.88 & 6.21 & \CC \textbf{21.78} & 11.29 & - & - & - \\
        & YAYI-UIE  & 5.11 & 8.21 & 19.74 & 11.02 & \CC \textbf{63.06} & \CB 59.42 & \CB 61.24  \\
        \cline{2-9}
        & Baichuan2-{\ours}  & 7.64 & 10.42 & \CB 20.40 & 12.82 & 57.93 & \CC \textbf{65.43} & \CC \textbf{61.68} \\
        & LLaMA2-{\ours} & \CC \textbf{12.55} & \CB 11.30 & 18.47 & 14.11 & 43.26 & 35.71 & 39.49 \\
        & Qwen1.5-{\ours} & 11.93 & 10.57 & 20.22 & \CB 14.24 & \CB 59.49 & 58.86 & 59.18 \\
        & LLaMA3-{\ours} & \CB 12.10 & 10.96 & 19.20 & 14.09 & 48.19 & 42.59 & 45.39 \\
        & OneKE & 11.88 & \CC \textbf{13.26} & 20.11 & \CC \textbf{15.08} & 58.83 & 62.38 & 60.61  \\
        \bottomrule 
    \end{tabular}
    \caption{Zero-shot performance on Event Extraction (EE) task. Within each column, \colorbox{aliceblue}{shadow} and \colorbox{babyblue}{shadow} represent the top 2 results.}
    \label{tab:zero-ee}
\end{table*}

%% file: tabs/supervised_ner.tex
\begin{table*}[t]
    \centering
    \begin{tabular}{l|cc|ccc}
    \toprule  
    Dataset & InstructUIE & YAYI-UIE & Baichuan2-{\ours} & LLaMA2-{\ours} & OneKE \\
    \midrule
    ACE2005 & \CC \textbf{86.66}  & 81.78 & 81.86 & 81.14 & \CB 83.45 \\
    AnatEM & \CC \textbf{90.89}  & 76.54 & 87.21 & 86.90 & \CB 87.88 \\
    BC2GM & \CC \textbf{85.16}  & 82.05 & 80.73 & \CB 83.07 & 82.05 \\
    BC4CHEMD & 90.30  & 88.46  & \CB 90.45 & 90.07 & \CC \textbf{90.56} \\
    BC5CDR & \CC \textbf{89.59}  & 83.67 & 88.07 &  88.01 & \CB 88.45 \\
    CoNLL2003 & 92.94  & \CC \textbf{96.77}  & 92.49 & 92.98 & \CB 93.04 \\
    FabNER & 76.20 & 72.63 & \CB 77.07 & 76.33 & \CC \textbf{81.06} \\
    FindVehicle & 89.47 & 98.47 & \CB 98.49 & 97.91 & \CC \textbf{99.45} \\
    GENIA-Ent & 74.71 & 75.21 & 76.66 &  \CB 77.32 & \CC \textbf{78.29} \\
    HarveyNER & \CC \textbf{88.79} & 69.57 & 67.70 & 62.64 & \CB 69.87 \\
    MIT Movie & 89.01 & 70.14 & 88.23 & \CB 89.54 & \CC \textbf{89.96} \\
    MIT Restaurant & \CC \textbf{82.55} & 79.38 & 79.85 & \CB 81.30 & 79.89 \\
    MultiNERD  & 92.32 & 88.42 & \CB 94.60 & 94.24 & \CC \textbf{94.69} \\
    NCBI-Disease & \CC \textbf{90.23} & 87.29 & 85.26 & \CB 87.59 & 86.95 \\
    Ontonotes & \CB 90.19 & 87.04 & 87.55 &  \CC \textbf{90.34} & 89.08 \\
    \textit{Avg} & \CC \textbf{87.27} & 82.49 & 85.08 & 85.29 & \CB 86.24 \\
    \hline
    MSRA  & - & \CC \textbf{95.57} & 87.99 & 86.32 & \CB 89.02 \\
    Resume NER  & - & -  & \CB 93.92 & 92.86 & \CC \textbf{95.84} \\
    CLUE NER  & - & - & \CC \textbf{80.19} & 76.57 & \CB 
 78.43 \\
    \bottomrule
    \end{tabular}
    \caption{Overall supervision results on  Named Entity Recognition (NER) datasets. Within each row, \colorbox{aliceblue}{shadow} and \colorbox{babyblue}{shadow} represent the top 2 results.}
\label{supervised-ner}
\end{table*}

%% file: tabs/supervised_re.tex
\begin{table*}[t]
    \centering
    \begin{tabular}{l|cc|ccc}
    \toprule  
    Dataset & InstructUIE & YAYI-UIE & Baichuan2-{\ours} & LLaMA2-{\ours} & OneKE \\
    \midrule
    ADE Corpus & 82.31 & 84.14 & 83.73 & \CB 85.87 & \CC \textbf{87.24} \\
    CoNLL2004 & \CB 78.48  & \CC \textbf{79.73} & 72.87 & 73.71 & 76.16 \\
    GIDS  & \CC \textbf{81.98} & 72.36 & 74.71 & 74.13 & \CB 76.69 \\
    KBP37  & 36.14  & 59.35 & \CB 65.09 & 61.49 & \CC \textbf{65.23} \\
    NYT &  90.47 & 89.97 & \CB 93.00 & 92.22 & \CC \textbf{94.04} \\
    NYT11-HRL & \CB 56.06 & \CC \textbf{57.53} & 53.19 & 54.86 & 55.56 \\
    SciERC & \CB 45.15 & 40.94 &  43.53 & 44.58 & \CC \textbf{45.89} \\
    Semeval-RE & \CC \textbf{73.23} & 61.02 & 58.47 & 57.61 & \CB 61.46 \\
    \textit{Avg} & 67.98 & \CB 68.13 & 68.07 & 68.06 & \CC \textbf{70.28} \\
    \hline
    CMeIE & - & - & \CB 49.16 & 47.40 & \CC \textbf{49.54} \\
    DuIE2.0 & - & \CC \textbf{81.19} & 75.61 & 74.34 & \CB 75.73 \\
    \bottomrule
    \end{tabular}
    \caption{Overall supervision results on Relation Extraction (RE) datasets. Within each row, \colorbox{aliceblue}{shadow} and \colorbox{babyblue}{shadow} represent the top 2 results.}
\label{supervised-re}
\end{table*}

%% file: tabs/supervised_ee.tex
\begin{table*}[t] 
    \centering
    \setlength{\tabcolsep}{3pt}
    \begin{tabular}{l|l|cc|ccc} 
    \toprule
        ~ & Dataset & InstructUIE & YAYI-UIE & Baichuan2-{\ours} & LLaMA-{\ours} & OneKE \\
        \hline
        \multirow{4}*{Trigger}  
        & ACE2005 & \CC \textbf{77.13} & 65.00 & \CB 72.46 & 70.63 & 71.17 \\
        & CASIE & \CC \textbf{67.80} & 63.00 & 60.07 & 61.27 & \CB 63.82 \\
        & PHEE & \CC \textbf{70.14} & 63.00 & 66.22 & 68.52  & \CB 68.60 \\
        ~ & \textit{Avg} & \CC \textbf{71.69} & 63.67 & 66.25 & 66.81 & \CB 67.86 \\
        \cline{2-7}
        & DuEE1.0  & - & 85.00 & \CC \textbf{86.73} & 84.01 & \CB 85.75 \\
        & DuEE-fin & - & 82.50 & \CC \textbf{83.54} & 79.00 & \CB 82.91 \\  
        \hline
        \multirow{4}*{Argument}   
        & ACE2005 & \CC \textbf{72.94} & 62.71 & \CB 63.90 & 62.69 & 62.75 \\
        & CASIE & \CB 63.53 & \CC \textbf{64.23} & 56.07 & 56.78 & 57.16 \\
        & PHEE & 62.91 & \CC \textbf{77.19} &  70.85 & 71.33 & \CB 72.84 \\
        ~ & \textit{Avg} & \CB 66.46 & \CC \textbf{68.04} & 63.60 & 63.61 & 64.25 \\
        \cline{2-7}
        & DuEE1.0  & - & \CC \textbf{79.08}  & \CB 75.63  & 73.79 & 75.40 \\
        & DuEE-fin & - & 70.02  & \CC \textbf{79.34}  & 73.08 & \CB 78.98 \\
        \bottomrule 
    \end{tabular}
    \caption{Overall supervision results on Event Extraction (EE) datasets. Within each row, \colorbox{aliceblue}{shadow} and \colorbox{babyblue}{shadow} represent the top 2 results.}
    \label{supervised-ee}
\end{table*}

%% file: tabs/statistic_ner.tex
\begin{table*}[htbp]
    \centering
    \setlength{\tabcolsep}{2.5pt}
    \begin{tabular}{l|l|c|cccc}
    \toprule  
    Task & Dataset & Domain & \#Schemas & \#Train & \#Val & \#Test \\
    \midrule
    \multirow{20}*{NER-en} 
    & AnatEM  ~\citep{DBLP:journals/bioinformatics/PyysaloA14} & Biomedical & 1 & 5667 & 2081 & 3758 \\
    & BC2GM ~\citep{DBLP:conf/icpr/KocamanT20}  & Biomedical& 1 & 12392 & 2483 & 4977 \\
    & BC4CHEMD~\citep{DBLP:conf/icpr/KocamanT20} & Biomedical & 1 & 30488 & 30468 & 26204 \\
    & NCBI-Disease ~\citep{DBLP:journals/jbi/DoganLL14} & Biomedical & 1 & 5432 & 923 & 940 \\
    & BC5CDR~\citep{DBLP:conf/iclr/00120GP23} & Biomedical & 2 & 4545 & 4569 & 4788 \\
    & HarveyNER ~\citep{DBLP:conf/naacl/ChenXZH22} & Social Media & 4 & 3553 & 1270 & 1260 \\
    & CoNLL2003  ~\citep{DBLP:conf/conll/SangM03} & News & 4 & 12613 & 3070 & 3184 \\
    & GENIA ~\citep{DBLP:conf/ismb/KimOTT03} & Biomedical & 5 & 14966 & 1657 & 1850 \\
    & ACE2005 ~\citep{ace2005-annotation}  & News  & 7 & 7134 & 964 & 1050 \\
    & MIT Restaurant ~\citep{DBLP:conf/icassp/LiuPCG13}  & Social Media & 8 & 7658 & - & 1520 \\
    & MIT Movie ~\citep{DBLP:conf/icassp/LiuPCG13} & Social Media & 12 & 9707 & - & 2441 \\
    & FabNER ~\citep{DBLP:journals/jim/KumarS22} & Scientific & 12 & 9421 & 2179 & 2064 \\
    & MultiNERD ~\citep{DBLP:conf/naacl/TedeschiN22} & Wikipedia & 16 & 130623 & 9994 & 9994 \\
    & Ontonotes ~\citep{DBLP:conf/naacl/PradhanX09} & General & 18 & 54994 & 7997 & 7782 \\
    & FindVehicle ~\citep{DBLP:journals/corr/abs-2304-10893} & Traffic  & 21 & 21547 & - & 20769 \\
    & CrossNER\_Politics$\dagger$ ~\citep{DBLP:conf/aaai/Liu0YDJCMF21} & Political & 9 & - & - & 650 \\
    & CrossNER\_Literature$\dagger$  ~\citep{DBLP:conf/aaai/Liu0YDJCMF21} & Literary & 12 & - & - & 416 \\
    & CrossNER\_Music$\dagger$  ~\citep{DBLP:conf/aaai/Liu0YDJCMF21} & Musical & 13 & - & - & 465 \\
    & CrossNER\_AI$\dagger$ ~\citep{DBLP:conf/aaai/Liu0YDJCMF21} & AI & 14 & - & - & 431 \\
    & CrossNER\_Science$\dagger$ ~\citep{DBLP:conf/aaai/Liu0YDJCMF21} & Scientific & 17 & - & - & 543 \\
    \midrule
    \multirow{5}*{NER-zh} 
    & MSRA NER ~\citep{DBLP:conf/acl-sighan/Levow06} & News & 3 & 40500 & 4500 & 3437 \\
    & Resume NER ~\citep{DBLP:conf/acl/ZhangY18} & Resume & 8 & 3799 & 463 & 476 \\
    & CLUE NER ~\citep{DBLP:journals/corr/abs-2001-04351} & News & 10 & 9674 & 1074 & 1343 \\
    & Weibo NER$\dagger$  ~\citep{DBLP:conf/emnlp/PengD15} & News & 4 & - & - & 258 \\
    & Boson$\dagger$ \ref{fn:boson} & News & 6 & - & - & 191 \\
    \bottomrule
    \end{tabular}
    \caption{Statistical data of Named Entity Recognition (NER) datasets, with an $\dagger$ indicating the zero-shot evaluation set not included in the training. CrossNER ~\citep{DBLP:conf/aaai/Liu0YDJCMF21} is divided into five subsets for our statistical analysis.}
    \label{statistic-ner}
\end{table*}

%% file: tabs/statistic_re.tex
\begin{table*}[htbp]
    \centering
    \setlength{\tabcolsep}{3.5pt}
    \begin{tabular}{l|l|c|cccc}
    \toprule  
    Task & Dataset & Domain & \#Schemas & \#Train & \#Val & \#Test \\
    \midrule
    \multirow{10}*{RE-en} 
    & ADE Corpus ~\citep{DBLP:journals/biomedsem/GurulingappaRT12} & Biomedical & 1 & 3416 & 427 & 428 \\
    & GIDS ~\citep{DBLP:conf/akbc/JatKT17} & News & 4 & 8525 & 1417 & 4307 \\
    & CoNLL2004 ~\citep{DBLP:conf/conll/CarrerasM04} & News & 5 & 922 & 231 & 288 \\
    & SciERC ~\citep{DBLP:conf/emnlp/LuanHOH18}  & Scientific & 7 & 1366 & 187 & 397 \\
    & Semeval-RE ~\citep{DBLP:conf/semeval/HendrickxKKNSPP10} & Scientific & 10 & 6478 & 1492 & 2714 \\
    & NYT11-HRL ~\citep{DBLP:conf/aaai/TakanobuZLH19} & News & 12 & 60765 & 146 & 362 \\
    & KBP37 ~\citep{DBLP:journals/corr/ZhangW15a} & News & 18 & 15911 & 1723 & 3405 \\
    & NYT ~\citep{DBLP:conf/pkdd/RiedelYM10} & News & 24 & 54412 & 4975 & 4985 \\
    & Wiki-ZSL ~\citep{DBLP:conf/naacl/ChenL21} $\dagger$ & Wikipedia & 83 & - & - & - \\
    & FewRel ~\citep{DBLP:conf/emnlp/HanZYWYLS18} $\dagger$  & Wikipedia & 100 & - & - & - \\
    \midrule
    \multirow{3}*{RE-zh} 
    & CMeIE  ~\citep{DBLP:conf/nlpcc/GuanZZXZ20} & Biomedical & 53 & 14339 & 3585 & - \\
    & DuIE2.0 ~\citep{DBLP:conf/nlpcc/LiHSJLJZLZ19} & News & 49 & 171126 & 20652 & - \\
    & COAE2016$\dagger$ \ref{fn:coae2016} & General & 9 & - & - & 971 \\
    & IPRE$\dagger$ ~\citep{wang2019ipre} & General & 35 & - & - & 3340 \\
    & SKE2020$\dagger$ \ref{fn:SKE2020} & News & 49 & - & - & 3601 \\
    \bottomrule
    \end{tabular}
    \caption{Statistical data of Relation Extraction (RE) datasets, with an $\dagger$ indicating the zero-shot evaluation set not included in the training. 
    The test sets for CMeIE and DuIE2.0 are not open-sourced, thus we use the validation sets as our evaluation set.
    For the FewRel and Wiki-ZSL datasets, we follow ~\citet{DBLP:conf/acl/ChiaBPS22}.
    }
    \label{statistic-re}
\end{table*}

%% file: tabs/statistic_ee.tex
\begin{table*}[htbp]
    \centering
    \begin{tabular}{l|l|c|cccc}
    \toprule  
    Task & Dataset & Domain  & \#Schemas & \#Train & \#Val & \#Test \\
    \midrule
    \multirow{3}*{EE-en} 
    & ACE2005 ~\citep{ace2005-annotation} & News & 33(22) & 3257 & 319 & 293 \\
    & CASIE ~\citep{DBLP:conf/aaai/SatyapanichFF20} & Cybersecurity & 5(26) & 3732 & 777 & 1492 \\
    & PHEE ~\citep{DBLP:conf/emnlp/Sun0PWJGK022} & Biomedical & 2(16) & 2897 & 960 & 968 \\
    & CrudeOilNews $\dagger$ ~\citep{DBLP:conf/lrec/LeeSSS22} & Oil News & 18(104) & - & - & 356 \\
    & RAMS $\dagger$ ~\citep{ebner-etal-2020-multi} & News & 106(398) & - & - & 887 \\
    & WikiEvents $\dagger$ ~\citep{DBLP:conf/naacl/LiJH21}  & Wikipedia & 31(81) & - & - & 249 \\
    \midrule
    \multirow{3}*{EE-zh} 
    & DuEE1.0 ~\citep{DBLP:conf/nlpcc/LiLPCPWLZ20} & News & 65(217) & 11908 & 1492 & - \\
    & DuEE-Fin ~\citep{DBLP:conf/nlpcc/HanZLXPZ22} & Finance & 13(91) & 7015 & 1171 & - \\
    & FewFC $\dagger$ ~\citep{DBLP:conf/aaai/Zhou0ZWXL21} & Finance & 5(29) & - & - & 2879 \\
    & CCF law $\dagger$\ref{fn:ccflaw} & Law & 9(39) & - & - & 971 \\
    \bottomrule
    \end{tabular}
    \caption{Statistical data of Event Extraction (EE) datasets, with an $\dagger$ indicating the zero-shot evaluation set not included in the training.
    The test sets for DuEE1.0 and DuEE-Fin are not open-sourced, thus we use the validation sets as our evaluation set.}
    \label{statistic-ee}
\end{table*}

%% file: tabs/instructions.tex
\begin{table*}[htbp]
    \centering
    \setlength{\tabcolsep}{2pt}
    \begin{tabular}{m{0.08\linewidth} m{0.9\linewidth}}
    \toprule    
    \textbf{Task} & \textbf{Instruction \& Output} \\ \midrule
    NER & \begin{lstlisting}
{
    "instruction": "You are an expert in named entity recognition. Please extract entities that match the schema definition from the input. Return an empty list if the entity type does not exist. Please respond in the format of a JSON string.",
    "schema": ["location", "else", "organization", "person"],
    "input": "The objective of the Basic Course on War is to provide for combatants of the EPR basic military knowledge for the armed conflict against the police and military apparatus of the bourgeoisie."
}
output = {
    "location": [], 
    "else": [], 
    "organization": ["EPR"], 
    "person": []
}

\end{lstlisting} \\  
    \midrule
    RE & \begin{lstlisting}
{
    "instruction": "You are an expert in relationship extraction. Please extract relationship triples that match the schema definition from the input. Return an empty list for relationships that do not exist. Please respond in the format of a JSON string.", 
    "schema": ["place of birth", "country capital", "country of administrative divisions", "company"], 
    "input": "Born on May 1 , 1927 , in Brichevo , Bessarabia in the present-day Republic of Moldova , Mr. Bertini emigrated to Palestine with his family as a child and pursued musical studies there , in Milan , and in Paris , where he worked with Nadia Boulanger and Arthur Honegger."
}
output = {
    "place of birth": [{"head": "Mr. Bertini", "tail": "Paris"}], 
    "country capital": [], 
    "country of administrative divisions": [], 
    "company": []
}

\end{lstlisting} \\
    \midrule
    EE & \begin{lstlisting}
{
    "instruction": "You are an expert in event extraction. Please extract events from the input that conform to the schema definition. Return an empty list for events that do not exist, and return NAN for arguments that do not exist. If an argument has multiple values, please return a list. Respond in the format of a JSON string.", 
    "schema": [{"event_type": "pardon", "trigger": true, "arguments": ["defendant"]}, {"event_type": "extradite", "trigger": true, "arguments": ["person", "agent", "destination", "origin"]}, {"event_type": "sue", "trigger": true, "arguments": ["place", "plaintiff"]}, {"event_type": "start position", "trigger": true, "arguments": ["person", "entity", "place"]}], 
    "input": "Ethical and legal issues in hiring Marinello"
}
output = {
    "pardon": [], 
    "extradite": [], 
    "sue": [], 
    "start position": [{"trigger": "hiring", "arguments": {"person": "Marinello", "entity": "NAN", "place": "NAN"}}]
}
\end{lstlisting} \\
    \bottomrule
    \end{tabular}
    \caption{Instructions and outputs for 3 tasks: Named Entity Recognition (NER), Relation Extraction (RE), and Event Extraction (EE).
    The instruction and output formats for {\ours} adopt a structure similar to JSON strings.}
    \label{tab:format}
\end{table*}